\documentclass[runningheads]{llncs}

\usepackage{graphicx}
\usepackage{amsmath}
\usepackage{times}
\usepackage{booktabs}
\usepackage{subcaption}

\usepackage[hidelinks]{hyperref}
\usepackage{gensymb}
\usepackage{enumitem}
\usepackage{tabularx}
\usepackage{url}
\usepackage{cite}
\usepackage{graphicx}
\usepackage[para]{footmisc}

\begin{document}

\title{Large Scale Audio-Visual Video Analytics Platform for Forensic Investigations of Terroristic Attacks}

\titlerunning{Large Scale Audio-Visual Video Analytics Platform}

\author{Alexander Schindler\inst{1} \and
        Martin Boyer\inst{1} \and
        Andrew Lindley\inst{1} \and
        David Schreiber\inst{1} \and
        Thomas Philipp\inst{2}}

\authorrunning{A. Schindler, M. Boyer, A. Lindley, D. Schreiber, T. Philipp}

\institute{Center for Digital Safety and Security, AIT Austrian Institute of Technology GmbH, Vienna 1210, Austria, \url{http://ait.ac.at} \\
\email{alexander.schindler@ait.ac.at} \and
LIquA - Linzer Institut f{\"u}r qualitative Analysen, Linz 4020, Austria, \url{http://liqua.net}\\
\email{thomas.philipp@liqua.net}}

\maketitle              % typeset the header of the contribution

%%%%%%%%%%%%%%%%%%%%%%%%%%%%%%%%%%%%%%%%%%%%%%%%%%%%%%%%%%%%%%%%%%%%%%%%%%%%%%%%
\begin{abstract}
%%%%%%%%%%%%%%%%%%%%%%%%%%%%%%%%%%%%%%%%%%%%%%%%%%%%%%%%%%%%%%%%%%%%%%%%%%%%%%%%

The forensic investigation of a terrorist attack poses a huge challenge to the investigative authorities, as several thousand hours of video footage need to be spotted. To assist law enforcement agencies (LEA) in identifying suspects and securing evidences, we present a platform which fuses information of surveillance cameras and video uploads from eyewitnesses. The platform integrates analytical modules for different input-modalities on a scalable architecture. Videos are analyzed according their acoustic and visual content. Specifically, Audio Event Detection is applied to index the content according to attack-specific acoustic concepts. Audio similarity search is utilized to identify similar video sequences recorded from different perspectives. Visual object detection and tracking are used to index the content according to relevant concepts. The heterogeneous results of the analytical modules are fused into a distributed index of visual and acoustic concepts to facilitate rapid start of investigations, following traits and investigating witness reports.

\end{abstract}

%%%%%%%%%%%%%%%%%%%%%%%%%%%%%%%%%%%%%%%%%%%%%%%%%%%%%%%%%%%%%%%%%%%%%%%%%%%%%%%%
\section{INTRODUCTION}
%%%%%%%%%%%%%%%%%%%%%%%%%%%%%%%%%%%%%%%%%%%%%%%%%%%%%%%%%%%%%%%%%%%%%%%%%%%%%%%%

% --- why was this system created?
The presented platform is a result of the project \textit{Flexible, semi-automatic Analysis System for the Evaluation of Mass Video Data (FLORIDA)} and is further developed in the project \textit{VICTORIA}. The aim of these projects is to facilitate the the work of investigators after a terrorist attack. In such events investigating video data is a major resource to spot suspects and to follow hints by civilian witnesses. From past attacks it is known that confiscated and publicly provided video content can sum up to thousands of hours (e.g. more than 5.000 hours at the \textit{Boston marathon bombing} attack).
%
% --- what are the biggest problems to solve?
Being able to promptly analyze mass video data with regards to content is increasingly important for complex investigative procedures, especially for those dealing with crime scenes. Currently, this data is analyzed manually which requires hundreds or thousands of hours of investigative work. As a result, extraction of first clues from videos after an attack takes a long time. Additionally, law enforcement agencies (LEA) may not be able to process all the videos, leaving important evidence and clues unnoticed. This effort continues to increase when evidence videos of civilian witnesses are uploaded multiple times. The prompt analysis of video data, however, is fundamental – especially in the event of terrorist attacks – to prevent immediate, subsequent attacks.
%
% --- how do we approach this?
The goal of this platform is to provide legally compliant tools for LEAs that will increase their effectiveness in analyzing mass video data and speed up investigative work. 
These tools include modules for acoustic and visual analysis of the video content, where especially the audio analysis tools provide a fast entry point to an investigation because most terroristic attacks emit characteristic sound events. An investigator can start viewing videos at such events and then progress forward or backward to identify suspects and evidences.

%A range of new methods for the analysis of video data will be developed in close cooperation with end users. These include geo-spatial crime scene reconstruction, interactive exploration of extracted information (Visual Analytics), prioritizing of video data, generic object search and audio analysis. Combining these methods will greatly reduce the amount of data to be analyzed by investigators, thus increasing focus on relevant events and supporting decision making.

% --- remainder
The remainder of this work is structured as follows: Section \ref{sec:rel_work} provides an overview of related work, Section \ref{sec:audio} details the audio analysis, Section \ref{sec:video} the video analysis module and Section \ref{sec:platform} the scalable platform. Section \ref{sec:ethics} summarizes the accompanying ethical research before we provide conclusions and an outlook to future work in Section \ref{sec:conclusions}.

%%%%%%%%%%%%%%%%%%%%%%%%%%%%%%%%%%%%%%%%%%%%%%%%%%%%%%%%%%%%%%%%%%%%%%%%%%%%%%%%
\section{Related Work}
\label{sec:rel_work}
% Related work section should be approx. 1 page
%%%%%%%%%%%%%%%%%%%%%%%%%%%%%%%%%%%%%%%%%%%%%%%%%%%%%%%%%%%%%%%%%%%%%%%%%%%%%%%%

\begin{itemize}[noitemsep, topsep=5pt, leftmargin=0pt]
\itemsep0.2em \renewcommand\labelitemi{}

% =====================================================================
\item \textbf{Audio Analysis:}
% =====================================================================
%
The audio analysis methods of the presented platform include modules for Audio Event Detection and Audio Similarity Retrieval. 
\textit{Audio Event Detection (AED)} systems combine detection and classification of acoustic concepts. Developments in this field have recently been driven by the annual international evaluation campaign \textit{Detection and Classification of Acoustic Scenes and Events} and its associated workshop \cite{giannoulis2013detection}. Most recent AED approaches are based on deep convolutional neural networks \cite{Lidy2016}, or recurrent convolutional neural networks \cite{Adavanne2016} which can also be efficiently trained on weakly labeled data \cite{Kukanov2017}. Such an approach is also taken for the AED module described in Section \ref{sec:audio}.
\textit{Audio Similarity} has been extensively studied especially in the research field of Music Information Retrieval (MIR) \cite{knees2016music}. Similarity estimations are generally based on extracting audio features from the audio signal and calculating feature variations using a metric function \cite{pampalk2005improvements}. A similar approach is followed in Section \ref{sec:audio}. Recent attempts to learn audio embeddings and similarity functions with neural networks has shown promising results \cite{kim2018one}.

% =====================================================================
\item \textbf{Video Analysis}
% =====================================================================
%
Video analytics software makes surveillance systems more efficient, by reducing the workload on security and management authorities. 
%Video Analytics solutions employ sophisticated computer vision algorithms, aimed at replicating human abilities, including tasks such as object classification, detection and tracking. 
%Object classification and detection means reporting which object is present in a video frame, annotating the image with bounding boxes around each object, and classifying each object, namely labeling each bounding box with the category (class) the object belongs to. Video tracking is the process of following a specific object over time, by associating target objects in consecutive video frames. Multi-target tracking is even more challenging due to the need to track simultaneously many targets without confusing their identities, and overcoming mutual occlusions. Finally, multi-class multi-target tracking denotes simultaneously tracking many objects from various class categories. 
Computer vision problems such as image classification, object detection and object tracking have traditionally been approached using hand-engineered features and machine learning algorithms design, both of which were largely independent \cite{Srinivas2016}. Over the last recent years, Deep Learning methods have been shown to outperform previous state-of-the-art machine learning techniques, with computer vision one of the most prominent cases \cite{voulodimos2018DeepLearning}. In contrast to previous approaches, deep neural networks (DNN) learn automatically the features required for tasks such as object detection and tracking.
%Moreover, these features were shown to outperform human-engineered features.  % citation missing
Among the various network architectures that were discovered and employed for computer vision tasks, the convolutional neural network (CNN) and recurrent neural network (RNN) were found to be best suitable for object classification, detection and tracking \cite{Srinivas2016,voulodimos2018DeepLearning}.

% =====================================================================
\item \textbf{Large Scale Workflow Management and Information Fusion:}
% =====================================================================
%
%Workflows \& Automation refers to software-based workflow management and process orchestration systems used for designing, structuring and executing interdependent business process tasks. 
A comprehensive overview of early work-flow management (WfMS) and business process management (BPM) systems is provided by \cite{Xu2011}. BPM is generally concerned with describing and controlling the flow of inter-dependent tasks whereas WfMSs aim at facilitating fully automated data flow oriented work-flows which can be described through a Directed Acyclic Graph (DAG). 
%In a DAG which are nodes represent the processing steps and edges the run-after dependencies. %, in the following called DAG workflow systems. Human interaction is not excluded, but usually plays a minor role. 
State-of-the-art implementations of scientific DAG work-flow systems, designed to run computationally intensive tasks on large, complex and heterogeneous data, are \textit{Taverna} \footnote{https://taverna.incubator.apache.org/}, \textit{Triana}  \footnote{http://www.trianacode.org/}, or \textit{Kepler}  \footnote{https://kepler-project.org/}. 
%Those systems are designed to run computationally intensive tasks on large, complex and heterogeneous data. 
Popular systems are \textit{Pegasus} \footnote{https://pegasus.isi.edu/} due to its direct relation to Grid- and Cloud Computing, \textit{Kepler} due its \textit{Hadoop} integration as well as \textit{Hadoop} itself. Recent developments in this area are domain specific languages, such as the functional language Cuneiform \cite{Brandt2015CuneiformAF} offering deep integration with Apache \textit{Hadoop} and a high flexibility in connecting with external environments. Further frameworks (languages and execution engine) derived from a Big Data context are \textit{Pig Latin} \cite{Olston2008} (part of the \textit{Hadoop} Ecosystem) and Apache Spark \cite{Zaharia2010} (in-memory processing framework). They are widely used in the scientific and commercial context to create work-flows for processing large data sets, but they are general purpose data analysis frameworks rather than specifically built to model work-flows. 
%
%DAG workflow systems have been developed in all kinds of other application domains as well. Their functionality varies significantly with respect to the set of requirements to which they respond. 
%The following list summarizes high level requirements that are of major importance specifically regarding audio-visual content management and archiving workflow execution: Workflow description language, Flow control and conditionals, distributed task execution, reliability and fault tolerance, Hadoop integration, extensibility and integration, planning and scheduling, monitoring and visualization \cite{Nadarajan2006SemanticBasedWC} \cite{salah2014}
%A number of open source DAG frameworks have been developed and released as open source.
Table \ref{table:DAG_workflow_systems} provides an overview on selected open source DAG frameworks, classified by high level requirements that are of major importance specifically regarding audio-visual content management and archiving \cite{Nadarajan2006SemanticBasedWC}.
%which are of particular relevance in relation to the requirements listed above.
%
The fusion of heterogeneous sensor data and multi-modal analytical results is still underrepresented in literature. A system combining results from various visual-analytical components for combined visualization is presented in \cite{Fan2017}. 

\end{itemize}

\setlength\intextsep{2mm}

%TODO fix footnote refs in table are broken
% Please add the following required packages to your document preamble:
% \usepackage{graphicx}
\begin{table}
\resizebox{\textwidth}{!}{%
\begin{tabular}{lcccccccc}
                                         & \textbf{Airflow}\footnotemark & \textbf{Mistral{\footnotemark}} & \textbf{Score{\footnotemark}} & \textbf{Spiff{\footnotemark}}                                        & \textbf{Oozie{\footnotemark}} & \textbf{Pinball{\footnotemark}}                                        & \textbf{Azkaban{\footnotemark}}                                                    & \textbf{Luigi{\footnotemark}}                                      \\
\textbf{Workflow description language}   & Python, Jinja           & YAML DSL                & YAML                  & \begin{tabular}[c]{@{}c@{}}XML, JSON, \\ Python\end{tabular} & XML                   & Python                                                         & \begin{tabular}[c]{@{}c@{}}Built-in Job types, \\ custom jobs\end{tabular} & Python                                                     \\
\textbf{Flow control and conditionals}   & NO                      & YES                     & YES                   & YES                                                          & NO                    & Minimum                                                        & NO                                                                         & NO                                                         \\
\textbf{Distributed task execution}      & YES                     & YES                     & YES                   & NO                                                           & NO                    & NO                                                             & NO                                                                         & NO                                                         \\
\textbf{Reliability and fault tolerance} & YES                     & YES                     & YES                   & NO                                                           & YES                   & YES                                                            & YES                                                                        & YES                                                        \\
\textbf{Hadoop integration}              & NO                      & NO                      & NO                    & NO                                                           & YES                   & YES                                                            & YES                                                                        & YES                                                        \\
\textbf{Extensibility and integration}   & Utilities               & Python                  & Python, Java          & NO                                                           & NO                    & \begin{tabular}[c]{@{}c@{}}Pluggable \\ Templates\end{tabular} & Plugins                                                                    & \begin{tabular}[c]{@{}c@{}}CLI \\ integration\end{tabular} \\
\textbf{Planning and Scheduling}         & YES                     & YES                     & NO                    & NO                                                           & YES                   & YES                                                            & YES                                                                        & NO                                                         \\
\textbf{Monitoring and visualization}    & Web-UI                  & NO                      & NO                    & NO                                                           & Web-UI                & Web-UI                                                         & Web-UI                                                                     & WF Graph Visualizer                                       
\end{tabular}%
}
\caption{Overview of open source DAG work-flow systems}
\label{table:DAG_workflow_systems}

\end{table}
\footnotetext{https://github.com/airbnb/airflow}
\footnotetext{https://wiki.openstack.org/wiki/Mistral}
\footnotetext{https://github.com/CloudSlang/score}
\footnotetext{https://github.com/knipknap/SpiffWorkflow/wiki}
\footnotetext{http://oozie.apache.org}
\footnotetext{https://github.com/pinterest/pinball}
\footnotetext{https://azkaban.github.io}
\footnotetext{https://github.com/spotify/luigi }

\vspace{-1em}

%%%%%%%%%%%%%%%%%%%%%%%%%%%%%%%%%%%%%%%%%%%%%%%%%%%%%%%%%%%%%%%%%%%%%%%%%%%%%%%%
\section{Audio Analysis}
\label{sec:audio}
%%%%%%%%%%%%%%%%%%%%%%%%%%%%%%%%%%%%%%%%%%%%%%%%%%%%%%%%%%%%%%%%%%%%%%%%%%%%%%%%

Audio analysis is one of the key components of this platform. Due to the destructive intention of a terroristic act, this often emits one or more loud acoustic events which are captured from microphones disregarding the direction the sound originates from. Besides the higher perceptive field of acoustic information, many relevant events are non-visual or happen too fast to be captured by standard cameras (e.g. alarms, screams, gunshots). Thus, we apply audio analysis to index the video content according audible events and to provide an entry point for the investigations.

% === BEGIN - FIGURE ===========================
\begin{figure*}[t] 
\centering
\includegraphics[width=1.0\textwidth]{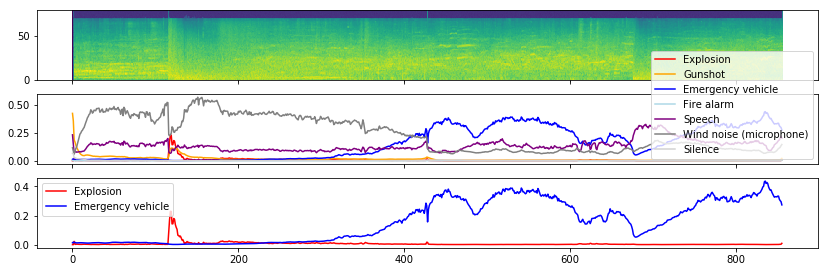}
\caption{Audio Event Detection example result (Bombing at Boston Marathon 2013). Top chart: Log-scaled Mel-Spectrogram of the audio signal. Middle chart: Probabilities for different acoustic events. Bottom chart: Explosion of the bomb on the left side, arrival of emergency vehicles from the center to the right of the chart.\vspace{-1em}}
\label{fig:audio_analysis_results}
\end{figure*}
% === END - FIGURE =============================

\begin{figure*}[t] 
\includegraphics[width=1.0\textwidth]{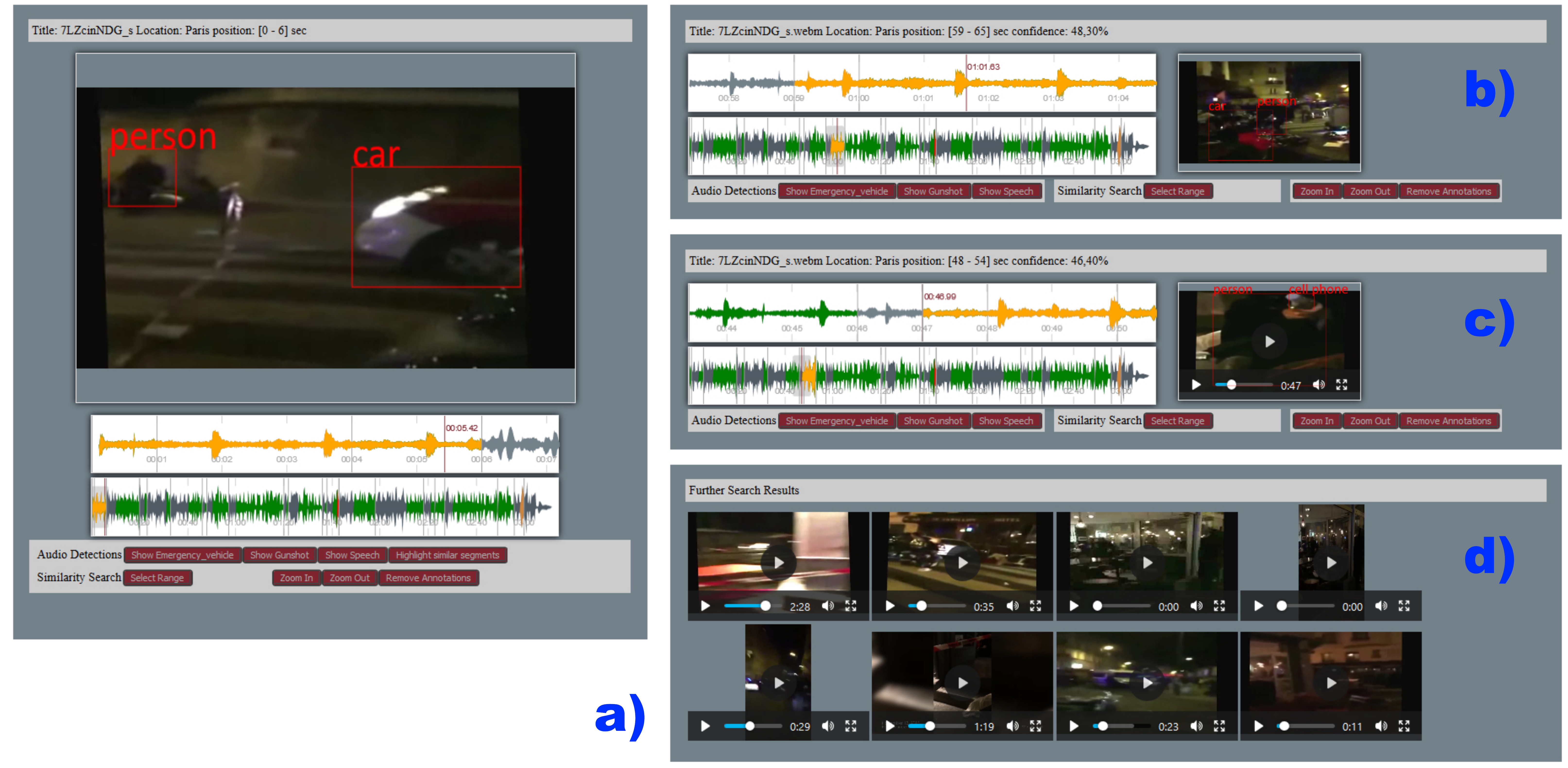}
\caption{\textbf{Audio-visual analysis results example}: a) reference video with detected audio events \textit{Gunshots} (green) and visual objects \textit{Person}, \textit{Car} (red bounding-box) and segment selected for similarity search (orange). b) video containing most similar audio sequence (orange) c) Second most similar sounding video segment (orange). d) further relevant videos ranked by audio similarity.\vspace{-1.5em}}
\label{fig:analysis_results}
\end{figure*}

\begin{itemize}[noitemsep, topsep=5pt, leftmargin=0pt]
\itemsep0.3em \renewcommand\labelitemi{}

% -------------------------------------------------------------------
\item \textbf{Audio Event Detection:}
The Audio Event Detection (AED) and recognition module is intended to be one of the primary entry points for investigations. Reports by civilian witnesses often refer to acoustic events (e.g. 'there was a loud noise and then something happened'). By indexing loud noises such as explosions, investigators can immediately pre-select videos where explosions are detected. This can be extended to the type of weapon used in the attack such as gunshots emitted by firearms and horns by trucks. The developed audio event detection and recognition method is based on deep neural networks. More specifically, the approach is a combination of the models we have developed and successfully evaluated in the \textit{Detection and Classification of Acoustic Scenes and Events (DCASE)} \cite{mesaros2016tut} international evaluation campaign \cite{Lidy2016,schindler2016comparing,schindler2017multi}, and the approach presented in \cite{xu2017attention}. The applied model uses Recurrent Convolutional Neural Networks with an attention layer.
%
% pre-processing
In a first step, the audio signal is extracted from the video containers, decoded and re-sampled to 44.100Hz single channel audio. 437.588 samples (9.92 seconds) are used as input, which are transformed to log-scaled Mel-Spectrograms, using 80 Mel-bands and a Short-Term Fourier transformed (STFT)  window size of 2048 samples with 1024 samples hop length. This preprocessing is directly performed on the GPU using the \textit{Kapre} signal processing layer \cite{choi2017kapre}.
% model
The normalized, decibel transformed input $ I $ is processed by a rectified linear convolution layer with 240 filter kernels of shape 30x1. Using global average pooling on the feature maps, audio embeddings with 240 dimensions are learned. This transformation is applied sequentially along the temporal axis of the Mel-Spectrogram, resulting in 428 audio embeddings (one for each STFT window). This 428x240 embedding space $ E $ is used as input for a stack of three bi-directional Gated Recurrent Units (GRU) \cite{chung2014empirical} followed by a rectified linear fully connected layer as well as a sigmoid fully connected layer with the number of units corresponding to the number of the to be predicted classes. A sigmoid scaled attention layer was further applied to each input frame of $ I $ which was multiplied with $ E $ as well as with the final prediction f the model. The final output of the model are probabilities for the presence for each of nine predefined sound events including \textit{Gunshot, Explosion, Speech, Emergency vehicle} and \textit{Fire Alarm} (see Figure \ref{fig:analysis_results}a).
The model was trained on a preprocessed subset of the \textit{Audioset} dataset \cite{gemmeke2017audio}. Preprocessing contained flattening of ontological hierarchies, resolving semantic overlaps, removing out-of-context classes (e.g. Music), re-grouping of classes and a final selection of task-relevant classes. 

% -------------------------------------------------------------------
\item \textbf{Audio Similarity Search:}
Indexing videos according to predefined categories provides a fast way to start an investigation but it is limited by the type and number of classes defined and undefined events such as \textit{train passing} cannot be detected. To overcome this obstacle and to facilitate the search for any acoustic pattern, an acoustic similarity function is added to search for videos with similar audio content. The approach to estimate the audio similarity is based on \cite{schindler2016europeana} where audio features are extracted, including \textit{Statistical Spectrum Descriptors} and \textit{Rhythm Patterns} \cite{lidy2007improving}, and distances are calculated between all extracted features using late fusion to merge the results. For this system, these features are extracted for each 6 seconds of audio content of every video file and the distances are calculated between all these features, facilitating a sub-segment similarity search. Further differences to \cite{schindler2016europeana} include omitting normalization by grouping features by their unit as well as using correlation distance for the \textit{Rhythm Patterns} feature-set, which showed better performance in preceding experiments. The audio similarity search serves several goals. First, if a suspect cannot be identified in a certain video, this function can be applied to identify video segments with similar acoustic signatures such as an emergency vehicle passing by, but any other sequence of sounds could be significant as well. Further, the recorded audio signal can be used for instant localization. Similar sound patterns have been recorded in near proximity to the emitting sources and thus the result of a similarity search provides video results for a referred location (see Figure \ref{fig:analysis_results}a-c).

\end{itemize}

% === BEGIN - FIGURE ===========================
% === END - FIGURE =============================

\vspace{-1em}
%%%%%%%%%%%%%%%%%%%%%%%%%%%%%%%%%%%%%%%%%%%%%%%%%%%%%%%%%%%%%%%%%%%%%%%%%%%%%%%%
\section{Video Analysis}
\label{sec:video}
%%%%%%%%%%%%%%%%%%%%%%%%%%%%%%%%%%%%%%%%%%%%%%%%%%%%%%%%%%%%%%%%%%%%%%%%%%%%%%%%

% -------------------------------------------------------------------

\begin{figure*}[t] 
\centering
\includegraphics[width=1.0\textwidth]{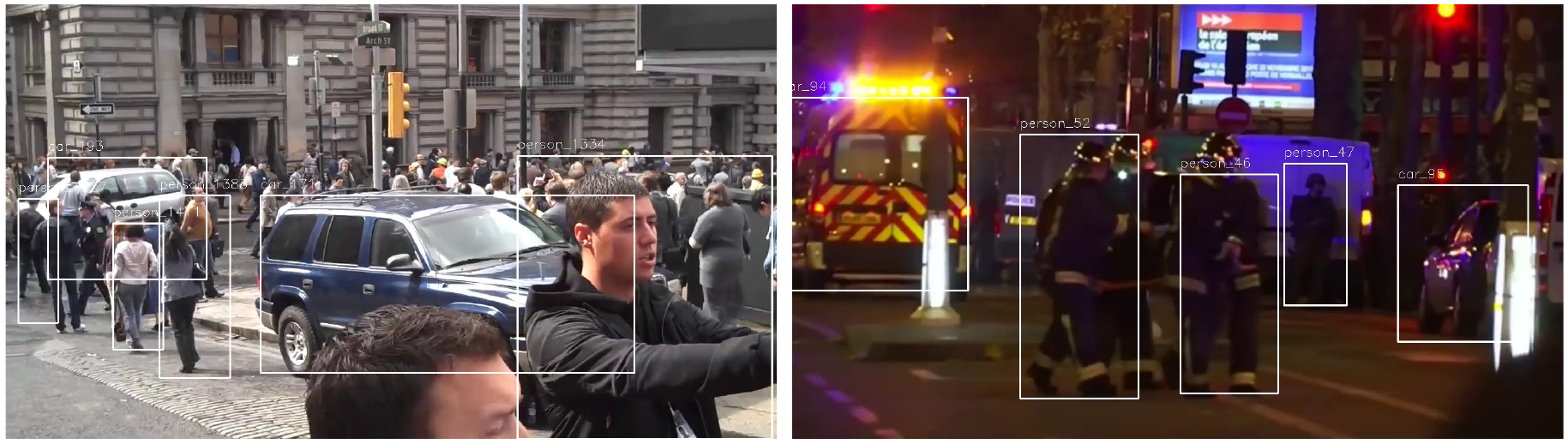}
\caption{Exemplary visual detection/tracking results for person and car classes in various typical criminal/terror scenes)\vspace{-1em}.}
\label{fig:visual_analysis_results}
\end{figure*}

\begin{itemize}[noitemsep, topsep=0pt, leftmargin=0pt]
\itemsep0.3em \renewcommand\labelitemi{}

\item \textbf{Generic Object Detection and Classification}
Object detection and classification identifies semantic concepts in video frames, including segmentation of the identified regions with bounding boxes and their labeling with the classified category such as \textit{car} or \textit{person}. This enables fast search queries that help identify specific scene content and therefore reduce the workload on law enforcement authorities.
%Object detection and classification is a core problem in computer vision. 
%Detection pipelines include the extraction of a set of robust features from input images and the application of classifiers to identify objects in the feature space. These classifiers are run either in sliding window fashion over the whole image or on some subset of regions in the image. 
In recent years, Deep Neural Networks (DNNs) have shown outstanding performance on image detection and classification tasks, replacing disparate parts such as feature extraction, by learning semantic representations and classifiers directly from input data, including the capability to learn more complex models than traditional approaches, and powerful object representations, without the need to hand design features \cite{voulodimos2018DeepLearning,redmon2017YOLO9000CVPR}.
YOLO (You Only Look Once) detector \cite{redmon2017YOLO9000CVPR} is one of the most popular CNN based detection algorithms, trained on over 9000 different object categories and real-time performance capacity \cite{redmon2017YOLO9000CVPR}. %Thus, it is suitable as a generic object detector. 
Evaluating different DNN based object detectors, we concluded that YOLO provides the best trade-off between accuracy and runtime behavior. 
The object detection module developed for the scalable forensic platform is based on the YOLO detector. It has been optimized to fit into the distributed environment and to store the results in the distributed database index. Figure \ref{fig:analysis_results} provides example outputs of this module.

\item \textbf{Multi-class Multi-target Tracking}
Visual tracking is a challenging task in computer vision due to target deformations, illumination variations, scale changes, fast and abrupt motion, partial occlusions, motion blur, and background clutter \cite{Ning2016}. The task of multi-target tracking consists of simultaneously detecting multiple targets at each time frame and matching their identities in different frames, yielding a set of target trajectories over time. Given a new frame, the tracker associates the already tracked targets with the newly detected objects (“tracking-by-detection” paradigm). 
Multi-target tracking is more challenging than the single target case, as interaction between targets and mutual occlusions of targets might cause identity switches between similar targets. 
%Most deep learning approaches are dedicated to single object tracking. 
There has been only little work related to multi-target tracking, presumably due to the following difficulties: First, %due to the large number of parameters, 
deep models require huge amounts of training data, which is not yet available in the case of multi-target tracking. Second, both the data and the desired solution can be quite variable. One is faced with both discrete (target labels) and continuous (position and scale) variables, unknown size of input and output, and variable lengths of video sequences \cite{Milan2016}. Finally, we note that no neural network based trackers were yet published which handle the general multi-class multi-target case.

DNN based multi-target trackers are trained either on appearance features \cite{Ning2016,Wojke2017,Milan2016}, or on some combination of appearance, motion and interaction features \cite{Sadeghian2017}. In  \cite{Wojke2017}, appearance-based association between targets and new objects is learned by CNNs, based on single frames. More robustly, appearance features are first learned on single frames using CNNs, and then, the long-term temporal dependencies in the sequence of observations are learned by RNNs, more precisely by Long-Short-Term-Memory (LSTM) networks \cite{Ning2016,Sadeghian2017}. The jointly trained neural networks then compute the association probability for each tracked target and newly detected object. Finally, an optimization algorithm is used to find the optimal matching between targets and new objects \cite{Ning2016,Sadeghian2017}. 
The developed system is an approach to a real-time multi-class multi-target tracking method, trained and optimized on the specific object categories needed in a forensic crime-scene and post-attack scenario investigation. Currently we employ an appearance based tracker as in \cite{Wojke2017}. Additional work in progress aims to add additional features such as targets motion and mutual interaction \cite{Sadeghian2017}, as well as learning temporal dependencies as in \cite{Ning2016}\cite{Sadeghian2017}. Due to various difficulties mentioned earlier in this section, we currently do not attempt to build and train one network for multi-class multi-target tracking. Rather, for each class, a multi-target tracker as in \cite{Wojke2017} is trained separately. During runtime, one tracker instance per class is activated, where all trackers are running in parallel. Figure \ref{fig:visual_analysis_results}) shows exemplary detection/tracking results for person and car classes in various typical criminal/terror scenes.

% -------------------------------------------------------------------

\item \textbf{Integration within the Connected Vision framework}
The generic object detection and multi-class multi-target tracking methods are integrated in a novel framework developed by AIT, denoted as Connected Vision \cite{boyer2014CVPoster}, which provides a modular, service-oriented and scalable approach allowing to process computer vision tasks in a distributed manner. The objective of Connected Vision is to create a video computation toolbox for rapid development of computer vision applications. To solve a complex computer vision task, two or more modules are combined to build a module chain. In our case, the modules are (i) video import, (ii) generic object detector, and (iii) multiple instances of a multi-target tracker, one module for each class of objects. Each Connected Vision module is an autonomous web-service that communicates via a Representational State Transfer (REST) interface, collects data from multiple sources (e.g. real-world physical sensors or other modules’ outputs), processes the data according to its configuration, stores the results for later retrieval and provides them to multiple consumers. The communication protocol is designed to support live (e.g. network camera) as well as archived data (e.g. video file) to be processed.   

\end{itemize}

% === BEGIN - FIGURE ===========================

% === END - FIGURE =============================

% === BEGIN - FIGURE ===========================
%\begin{minipage}{.98\columnwidth}
%   \centering
%   \includegraphics[width=.98\columnwidth]{for_designers.pdf}
%   \captionof{figure}{\textbf{Connected Vision Module}}
%   \label{fig:connected_vision_module}
%\end{minipage}
% === END - FIGURE =============================

%%%%%%%%%%%%%%%%%%%%%%%%%%%%%%%%%%%%%%%%%%%%%%%%%%%%%%%%%%%%%%%%%%%%%%%%%%%%%%%%
\section{Scalable Analysis Platform}
\label{sec:platform}
%%%%%%%%%%%%%%%%%%%%%%%%%%%%%%%%%%%%%%%%%%%%%%%%%%%%%%%%%%%%%%%%%%%%%%%%%%%%%%%%

\begin{figure*}[t] 
\centering
\includegraphics[width=0.9\textwidth]{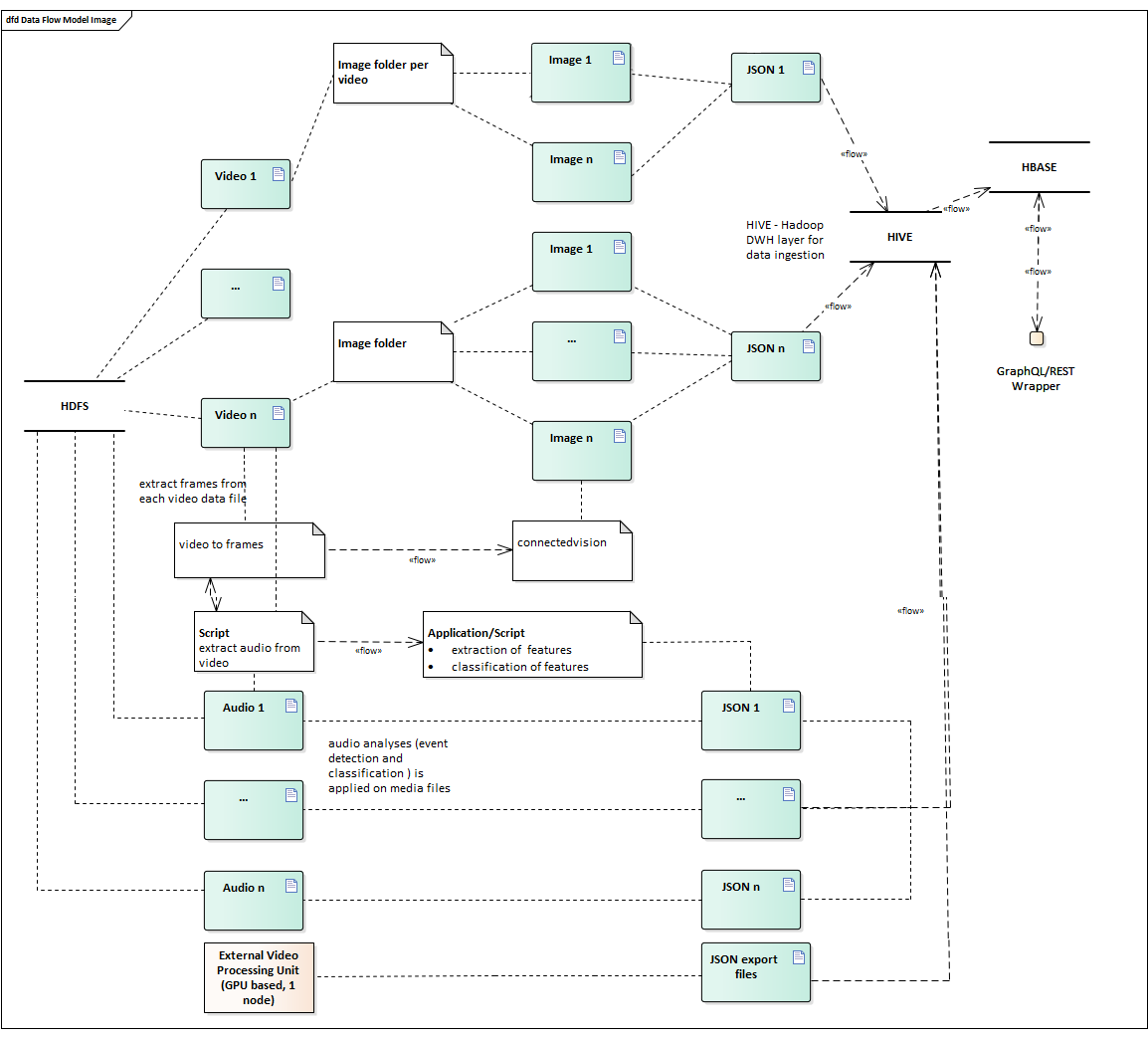}
\caption{Data Flow Model of the FLORIDA Scalable Analytics Platform}
\label{fig:florida_platform}
\end{figure*}

Two of the main goals of the developed platform are a) in the case of an attack scenario, it should be possible to probe the provided video media material (mass data) as quickly as possible and b) Law Enforcement Agency (LEAs) investigators should gain better insight by screening the most essential material from different perspective and by focusing on specific events through the help of an integrated Scalable Analysis Platform (SAP).
The FLORIDA SAP integrates the developed advanced analysis modules and performs the tasks \textit{video data ingestions}, \textit{data preparation and preprocessing}, \textit{feature extraction} and \textit{model fitting}. These fitted analysis models are then applied to the preprocessed video content. This is implemented on an Apache Hadoop\footnote{http://hadoop.apache.org/} platform. The analysis results are stored in an Apache HBASE\footnote{https://hbase.apache.org/} database with an GraphQL\footnote{https://graphql.org/} layer on top to provide dynamic access for the clients and the visualization of the calculated results.
The hardware cluster consists of seven compute nodes and a name node which acts as the orchestrator of the Cloudera Hadoop platform. The underlying commodity hardware (HW) consists of Dell R320 rack servers, Xeon® CPU, E5-2430, 2.2 GHz with 6 cores / 12 threads each and about 63 GB of available RAM and a storage capacity of 11TB of disk space (HDD) per data node. Server administration, cluster configuration, update management as well as software distribution on the nodes is automated via Ansible \footnote{Ansible https://www.ansible.com/ }. Tasks for rolling out the Zookeeper, Hive or HBASE configuration, distributing the /etc/hotsts files as well as installing the audio and video feature extraction tools and their software dependencies (such as Python scripts, Linux packages as ffmpeg and libraries as TensorFlow) are specified in the YAML syntax.
ToMaR \cite{Schmidt2014Tomar} is a generic MapReduce wrapper for third-party command-line and Java applications that were not originally designed for the usage in an HDFS environment. It supports tools that read input based on local file pointers or stdin/stdout streams uses control-files to specify individual jobs. The wrapper integrates the applications in such a way that the required files for execution on the respectively integrated worker nodes (nodes 0..5) are locally copied into the Hadoop cache from HDFS and the application can thus be executed in parallel to the number of active nodes and the results are then written back to HDFS. The individual components (ToMaR jobs, shell scripts for import of JSON data into and generation of HBASE tables via HIVE, etc.) are combined into master workflows using Apache Oozie Workflow Scheduler for Hadoop \footnote{ http://oozie.apache.org/} and event triggers. (see Figure \ref{fig:florida_platform}) This requires the setup of ZooKeeper for handling failover or orchestration of the components.  In order to execute Hive and HBase in a distributed environment, Hadoop must be executed in Fully Distributed Mode and requires the setup a so-called metastore for state information on Hive and Oozie for Hadoop. ZooKeeper is set up in the SAP as ZooKeeper ensemble on node node2, node5, and the master name node.

\vspace{-0.5em}
%%%%%%%%%%%%%%%%%%%%%%%%%%%%%%%%%%%%%%%%%%%%%%%%%%%%%%%%%%%%%%%%%%%%%%%%%%%%%%%%
\section{Accompanying ethical research}
\label{sec:ethics}
%%%%%%%%%%%%%%%%%%%%%%%%%%%%%%%%%%%%%%%%%%%%%%%%%%%%%%%%%%%%%%%%%%%%%%%%%%%%%%%%

The FLORIDA project is part of KIRAS, an Austrian research promotion program managed by the Austrian Research Promotion Agency (FFG), the national funding agency for industrial research and development in Austria. In KIRAS projects, an integrative approach is mandatory, which is based not only on technological solutions but also on a social science and humanities approach. For this reason, in FLORIDA accompanying ethical research are conducted.

\begin{itemize}[noitemsep, topsep=0pt, leftmargin=0pt]
\itemsep0.3em \renewcommand\labelitemi{}

\item \textbf{Ethics of security and ethics of technology: }
On the background of fundamental ethical principles, analyses in FLORIDA primarily focus on ethics of security and ethics of technology, but also include other ethical subdivisions, e.g. data, information, Internet and media ethics. In the context of security ethics questions about the 'right' or 'wrong' use of a certain security technology are reflected upon, for example by asking about the concept of security used, by linking security actions back to the question of 'good living' and by pointing out alternative options for actions that include comprehensive social and societal responsibility.\cite{rampp2014} By taking technical ethics into account the focus is directed to the ethical reflection of conditions, purposes, means and consequences of technology and scientific-technical progress.\cite{grunwald2013} This leads to questions relating to fundamental ethical principles such as security or freedom, but also to the choice, responsibility or compatibility of used technology: 'Is the chosen technology good?', 'Is the chosen technology safe?' or 'What are the consequences of using this technology?'

\item \textbf{Ethical criteria and questions for security systems: }
Based on these theoretical reflections and on knowledge from past civil security research projects such as THEBEN\footnote{THEBEN (Terahertz Detection Systems: Ethical Monitoring, Evaluation and Determination of Standards) was a research project 10-2007 to 12-2010) within the framework of the program for civil security research in Germany: https://www.uni-tuebingen.de/en/11265}, MuViT\footnote{MuViT (Mustererkennung und Video-Tracking) was a research project (10-2007 to 12-2010)  within the framework of the program for civil security research in Germany: http://www.uni-tuebingen.de/de/49647} or PARIS\footnote{PARIS (PrivAcy pReserving Infrastructure for Surveillance) was a research project (01-2013 to 02-2016) with partners from France, Belgium and Austria, funded by the 7th Framework Program for Research and Technological Development: https://www.paris-project.org/}, ethical criteria for security systems were defined and an ethical catalog with around 80 questions was compiled, which can be used in the funding, development and usage of security systems. During FLORIDA's accompanying research, the ethical criteria served as a framework for evaluating the planned use cases and the various development stages of the technological prototype. Some of the questions were the basis for qualitative interviews with potential users at Austrian police organization and intelligence agencies on the one hand and the technical developers of FLORIDA on the other. Here, for example, questions were asked about the ethical pre-understanding, the ethical risks in development or ethical responsibility, but also about specific technological challenges such as the prevention of a scenario creep, i.e. a step by step scenario extension of a system or parts of a system. For FLORIDA, this would mean to prevent individual functions that were only developed for a post-attack scenario from being expanded step by step later and finally being used for an observation scenario. On a technical level, this seems to be difficult or even impossible to prevent in general. Strategies that can be applied here rather focus on the basic handling of organizations or companies in the use of security-related systems, for example adequate authorization systems for the use of functions or the implementation of a security certification system that contains clear guidelines for possible functional extensions. FLORIDA, however, also includes some technological details that make a scenario creep at least more difficult. To give an example: In the audio analysis only predefined audio events are classified and implemented that can occur during attacks (e.g. shots, sirens or detonations) and the system is trained with them. This function of FLORIDA is therefore useful for a specific scenario like terror attacks, while it remains largely useless in other scenarios like an observation. Moreover, subsequent adaptations are difficult to carry out because complex and time-consuming learning processes are what make the functionality of audio analysis possible here.

\end{itemize}

%%%%%%%%%%%%%%%%%%%%%%%%%%%%%%%%%%%%%%%%%%%%%%%%%%%%%%%%%%%%%%%%%%%%%%%%%%%%%%%%
\section{Conclusions and Future Work}
\label{sec:conclusions}
%%%%%%%%%%%%%%%%%%%%%%%%%%%%%%%%%%%%%%%%%%%%%%%%%%%%%%%%%%%%%%%%%%%%%%%%%%%%%%%%

The described platform integrates audio-visual analysis modules on a scalable platform and enables a fast start and rapid progress of investigations with the audio event detection module serving as a fast entry point. From these events, investigators can navigate through the video content, by either using the visual tracking modules on identified persons or objects, or use the audio similarity search to find related video content, increasing the chances for identification. 

As part of future work we intend to focus on \textit{Audio Synchronization}, because time information provided with video meta-data can not be considered accurate if the capturing equipment is not synchronized with a unified time-server. Personal cameras for example commonly reset their internal clock to a hard-coded time-stamp after complete battery drain. To align video content with unreliable time information, audio synchronization is applied. Audio features sensitive to peaking audio events are applied to extract patterns which are significant for a recorded acoustic scene. These patterns are then matched by minimizing the difference of their feature values over sliding windows. To find clusters of mutually synchronous videos, audio similarity retrieval is combined with audio synchronization. Finally, mutual offsets are calculated between the videos of a cluster which are used to schedule synchronous playback of the videos.	\\

\noindent
\textit{\textbf{Acknowledgements} This article has been made possible partly by received funding from the European Union’s Horizon 2020 research and innovation program in the context of the VICTORIA project
under grant agreement no. SEC-740754 and the project FLORIDA, FFG Kooperative F\&E Projekte 2015, project no. 854768. }

%%%%%%%%%%%%%%%%%%%%%%%%%%%%%%%%%%%%%%%%%%%%%%%%%%%%%%%%%%%%%%%%%%%%%%%%%%%%%%%%
\bibliographystyle{unsrt}
\bibliography{ref}

\begin{thebibliography}{10}

\bibitem{giannoulis2013detection}
Dimitrios Giannoulis, Emmanouil Benetos, Dan Stowell, Mathias Rossignol,
  Mathieu Lagrange, and Mark~D Plumbley.
\newblock Detection and classification of acoustic scenes and events: An ieee
  aasp challenge.
\newblock In {\em Applications of Signal Processing to Audio and Acoustics
  (WASPAA), 2013 IEEE Workshop on}, pages 1--4. IEEE, 2013.

\bibitem{Lidy2016}
Thomas Lidy and Alexander Schindler.
\newblock {CQT}-based convolutional neural networks for audio scene
  classification.
\newblock In {\em Proceedings of the Detection and Classification of Acoustic
  Scenes and Events 2016 Workshop (DCASE2016)}, pages 60--64, September 2016.

\bibitem{Adavanne2016}
Sharath Adavanne, Giambattista Parascandolo, Pasi Pertilä, Toni Heittola, and
  Tuomas Virtanen.
\newblock Sound event detection in multichannel audio using spatial and
  harmonic features.
\newblock Technical report, DCASE2016 Challenge, September 2016.

\bibitem{Kukanov2017}
Ivan Kukanov, Ville Hautamäki, and Kong~Aik Lee.
\newblock Recurrent neural network and maximal figure of merit for acoustic
  event detection.
\newblock Technical report, DCASE2017 Challenge, 2017.

\bibitem{knees2016music}
Peter Knees and Markus Schedl.
\newblock {\em Music similarity and retrieval: an introduction to audio-and
  web-based strategies}, volume~36.
\newblock Springer, 2016.

\bibitem{pampalk2005improvements}
Elias Pampalk, Arthur Flexer, Gerhard Widmer, et~al.
\newblock Improvements of audio-based music similarity and genre classificaton.
\newblock In {\em ISMIR}, volume~5, pages 634--637. London, UK, 2005.

\bibitem{kim2018one}
Jaehun Kim, Juli{\'a}n Urbano, Cynthia Liem, and Alan Hanjalic.
\newblock One deep music representation to rule them all?: A comparative
  analysis of different representation learning strategies.
\newblock {\em arXiv preprint arXiv:1802.04051}, 2018.

\bibitem{Srinivas2016}
Suraj Srinivas, Ravi~Kiran Sarvadevabhatla, and Konda~Reddy Mopuri.
\newblock A taxonomy of deep convolutional neural netwprks for computer vision,
  2016.

\bibitem{voulodimos2018DeepLearning}
Athanasios Voulodimos, Nikolaos Doulamis, Anastasios Doulamis, and Eftychios
  Protopapadakis.
\newblock {Deep Learning for Computer Vision: A Brief Review}.
\newblock In {\em Hindawi Computational Intelligence and Neuroscience}, volume
  2018, pages 1--13.
\newblock { Article ID 7068349}.

\bibitem{Xu2011}
L.~D. Xu.
\newblock Enterprise systems: State-of-the-art and future trends.
\newblock {\em IEEE Transactions on Industrial Informatics}, 7(4):630--640, Nov
  2011.

\bibitem{Brandt2015CuneiformAF}
J{\"o}rgen Brandt, Marc Bux, and Ulf Leser.
\newblock Cuneiform: a functional language for large scale scientific data
  analysis.
\newblock In {\em EDBT/ICDT Workshops}, 2015.

\bibitem{Olston2008}
Christopher Olston, Benjamin Reed, Utkarsh Srivastava, Ravi Kumar, and Andrew
  Tomkins.
\newblock Pig latin: a not-so-foreign language for data processing.
\newblock In {\em Proc of int. conf. on Management of data (SIGMOD '08)}, pages
  1099--1110. ACM, 2008.

\bibitem{Zaharia2010}
Matei Zaharia, N.~M.~Mosharaf Chowdhury, Michael Franklin, Scott Shenker, and
  Ion Stoica.
\newblock Spark: Cluster computing with working sets.
\newblock Technical Report UCB/EECS-2010-53, EECS Department, University of
  California, Berkeley, May 2010.

\bibitem{Nadarajan2006SemanticBasedWC}
Gayathri Nadarajan, Y.-H. Chen-Burger, and James Malone.
\newblock Semantic-based workflow composition for video processing in the grid.
\newblock {\em 2006 IEEE/WIC/ACM International Conference on Web Intelligence
  (WI 2006 Main Conference Proceedings)(WI'06)}, pages 161--165, 2006.

\bibitem{Fan2017}
Ching~Tang Fan, Yuan~Kai Wang, and Cai~Ren Huang.
\newblock {Heterogeneous information fusion and visualization for a large-scale
  intelligent video surveillance system}.
\newblock {\em IEEE Transactions on Systems, Man, and Cybernetics: Systems},
  47(4):593--604, 2017.

\bibitem{mesaros2016tut}
Annamaria Mesaros, Toni Heittola, and Tuomas Virtanen.
\newblock Tut database for acoustic scene classification and sound event
  detection.
\newblock In {\em 24th Europ. Signal Proc Conf (EUSIPCO)}, 2016.

\bibitem{schindler2016comparing}
Alexander Schindler, Thomas Lidy, and Andreas Rauber.
\newblock Comparing shallow versus deep neural network architectures for
  automatic music genre classification.
\newblock In {\em 9th Forum Media Technology (FMT2016)}, volume 1734, pages
  17--21. CEUR, 2016.

\bibitem{schindler2017multi}
Alexander Schindler, Thomas Lidy, and Andreas Rauber.
\newblock Multi-temporal resolution convolutional neural networks for acoustic
  scene classification.
\newblock In {\em Detect. and Classific. of Acoustic Scenes and Events Workshop
  (DCASE2017), Munich, Germany}, 2017.

\bibitem{xu2017attention}
Yong Xu, Qiuqiang Kong, Qiang Huang, Wenwu Wang, and Mark~D Plumbley.
\newblock Attention and localization based on a deep convolutional recurrent
  model for weakly supervised audio tagging.
\newblock {\em arXiv preprint arXiv:1703.06052}, 2017.

\bibitem{choi2017kapre}
Keunwoo Choi, Deokjin Joo, and Juho Kim.
\newblock Kapre: On-gpu audio preprocessing layers for a quick implementation
  of deep neural network models with keras.
\newblock In {\em Machine Learning for Music Discovery Workshop at 34th Int.
  Conf. on Machine Learning}. ICML, 2017.

\bibitem{chung2014empirical}
Junyoung Chung, Caglar Gulcehre, KyungHyun Cho, and Yoshua Bengio.
\newblock Empirical evaluation of gated recurrent neural networks on sequence
  modeling.
\newblock {\em arXiv preprint arXiv:1412.3555}, 2014.

\bibitem{gemmeke2017audio}
Jort~F Gemmeke, Daniel~PW Ellis, Dylan Freedman, Aren Jansen, Wade Lawrence,
  R~Channing Moore, Manoj Plakal, and Marvin Ritter.
\newblock Audio set: An ontology and human-labeled dataset for audio events.
\newblock In {\em Acoustics, Speech and Signal Processing (ICASSP), 2017 IEEE
  International Conference on}, pages 776--780. IEEE, 2017.

\bibitem{schindler2016europeana}
Alexander Schindler, Sergiu Gordea, and Harry van Biessum.
\newblock The europeana sounds music information retrieval pilot.
\newblock In {\em Euro-Mediterranean Conference}, pages 109--117, 2016.

\bibitem{lidy2007improving}
Thomas Lidy, Andreas Rauber, Antonio Pertusa, and Jos{\'e} Manuel~I{\~n}esta
  Quereda.
\newblock Improving genre classification by combination of audio and symbolic
  descriptors using a transcription systems.
\newblock In {\em Proc. Int. Conf. Music Information Retrieval}, 2007.

\bibitem{redmon2017YOLO9000CVPR}
Joseph Redmon and Ali Farhadi.
\newblock {YOLO9000: Better, Faster, Stronger}.
\newblock In {\em Proceedings of the IEEE Conference on Computer Vision and
  Pattern Recognition (CVPR)}, July 2017.

\bibitem{Ning2016}
Guanghan Ning, Zhi Zhang, Chen Huang, Zhihai He, Xiaobo Ren, and Haohong Wang.
\newblock Spatially supervised recurrent convolutional neural networks for
  visual object tracking, 2016.

\bibitem{Milan2016}
Anton Milan, S.~Hamid Rezatofighi, Anthony Dick, Ian Reid, and Konrad
  Schindler.
\newblock Online multi-target tracking uing recurrent neural netwroks, 2016.

\bibitem{Wojke2017}
Nicolai Wojke, Alex Bewley, and Dietrich Paulus.
\newblock Simple online and realtime tracking with deep association metric,
  2017.

\bibitem{Sadeghian2017}
Nicolai Wojke, Alex Bewley, and Dietrich Paulus.
\newblock Tracking the untrackable: Learning to track multiple cues with
  long-term dependencies.
\newblock In {\em CVPR}, 2017.

\bibitem{boyer2014CVPoster}
Martin Boyer and Stephan Veigl.
\newblock A distributed system for secure, modular computer vision.
\newblock In {\em Future Security 2014 9th Future Security Security Research
  Conference, Berlin, September 16-18, 2014, Proceedings}, pages 696--699,
  Berlin, 2014.

\bibitem{Schmidt2014Tomar}
R.~Schmidt, M.~Rella, and S.~Schlarb.
\newblock Tomar -- a data generator for large volumes of content.
\newblock In {\em 14th IEEE/ACM Int. Symp. on Cluster, Cloud and Grid
  Computing}, pages 937--942, 2014.

\bibitem{rampp2014}
Benjamin Rampp.
\newblock {Zum Konzept der Sicherheit}.
\newblock In Regina {Ammicht Quinn}, editor, {\em Sicherheitsethik}, pages 51
  -- 61. Springer Fachmedien, Wiesbaden, 2014.

\bibitem{grunwald2013}
Armin Grunwald.
\newblock {Einleitung und {\"{U}}berblick}.
\newblock In Armin Grunwald, editor, {\em Handbuch Technikethik}, pages 1 --
  11. J.B. Metzler, Stuttgart, 2013.

\end{thebibliography}
%%%%%%%%%%%%%%%%%%%%%%%%%%%%%%%%%%%%%%%%%%%%%%%%%%%%%%%%%%%%%%%%%%%%%%%%%%%%%%%%

\end{document}